\documentclass[runningheads,a4paper]{llncs}

\usepackage{amssymb}
\usepackage{graphicx}
\usepackage{xcolor}

\usepackage{url}
\urldef{\mailsa}\path|m.grard@sileane.com|
\newcommand{\keywords}[1]{\par\addvspace\baselineskip
\noindent\keywordname\enspace\ignorespaces#1}

\usepackage{graphicx}
\usepackage{amsmath}
\usepackage{amssymb}
\usepackage{subfig}
\usepackage{xspace}
\usepackage{tabu}
\usepackage{multirow}
\usepackage{stmaryrd}
\usepackage{cite}
\makeatletter
\DeclareRobustCommand\onedot{\futurelet\@let@token\@onedot}
\newcommand*\@onedot{\ifx\@let@token.\else.\null\fi\xspace}
\newcommand*\eg{\emph{e.g}\onedot} 
\newcommand*\ie{\emph{i.e}\onedot} 
 
\newcommand*\etc{\emph{etc}\onedot}

\makeatother
\newlength\mywidth

\begin{document}

\mainmatter  

\title{Object segmentation in depth maps\\ with one user click and a synthetically\\ trained fully convolutional network}

\titlerunning{Object segmentation with one user click and a synthetically trained FCN}

%
%
\author{Matthieu Grard$^{1,2}$%
\and Romain Br\'egier$^{1,3}$
\and\\ Florian Sella$^1$
\and Emmanuel Dellandr\'ea$^2$
\and Liming Chen$^2$}
\authorrunning{Matthieu Grard}

\institute{$^1$Sil\'eane, 17 rue Descartes, F-42000 St \'Etienne, France\\
\mailsa\\
$^2$Universit\'e de Lyon, CNRS, \'Ecole Centrale de Lyon\\ LIRIS UMR5205, F-69134 Lyon, France\\
$^3$Univ. Grenoble Alpes, Inria, CNRS\\ Grenoble INP, LIG, F-38000 Grenoble, France\\}

%
%

\toctitle{Lecture Notes in Computer Science}
\tocauthor{Authors' Instructions}
\maketitle

\begin{abstract}
With more and more household objects built on plan\-ned obsolescence and consumed by a fast-growing population, hazardous waste recycling has become a critical challenge. Given the large variability of household waste, current recycling platforms mostly rely on human operators to analyze the scene, typically composed of many object instances piled up in bulk. Helping them by robotizing the unitary extraction is a key challenge to speed up this tedious process. Whereas supervised deep learning has proven very efficient for such object-level scene understanding, \eg, generic object detection and segmentation in everyday scenes, it however requires large sets of per-pixel labeled images, that are hardly available for numerous application contexts, including industrial robotics.

We thus propose a step towards a practical interactive application for generating an object-oriented robotic grasp, requiring as inputs only one depth map of the scene and one user click on the next object to extract. More precisely, we address in this paper the middle issue of object segmentation in top views of piles of bulk objects given a pixel location, namely seed, provided interactively by a human operator. We propose a two-fold framework for generating edge-driven instance segments. First, we repurpose a state-of-the-art fully convolutional object contour detector for seed-based instance segmentation by introducing the notion of \emph{edge-mask duality} with a novel patch-free and contour-oriented loss function. Second, we train one model using only synthetic scenes, instead of manually labeled training data. Our experimental results show that considering edge-mask duality for training an encoder-decoder network, as we suggest, outperforms a state-of-the-art patch-based network in the present application context.

\keywords{Interactive instance segmentation, supervised pixel-wise learning, synthetic training images}
\end{abstract}

\section{Introduction}
\label{intro}

Waste recycling is one of the main challenges lying ahead, as the ever-growing amount of obsolete household objects and industrial materials outstrips the deployed human resources. Current state legislations and policies thus entice industrialists to develop robotized waste sorting, which stands out as a competitive approach to protect humans from hazardous waste collection sites (toxic emissions, radioactive radiations, \etc). However, given the large variability of elements to sort, and the cluttered nature of waste, human expertise is hardly dispensable despite the great advances in machine learning. Towards a fast and accurate processing, enhancing human operators decisions with computer vision-guided robotic grasping seems therefore one of the current most suitable strategies. Numerous works on supervised learning for object-wise scene understanding in images of everyday scenes have indeed demonstrated remarkable results for real-time applications: \emph{generic object detection}, \ie, generating category-free object proposals, either as bounding box proposals \cite{FasterRCNN, HosangBDS15}, or as binary masks, namely segment proposals \cite{MaskRCNN, SharpMask, Pont-TusetABMM15, KrahenbuhlK15}; \emph{semantic segmentation}, \ie, assigning an instance-agnostic object category to each pixel \cite{Lin_2016_CVPR, LongSD15}; \emph{object contour detection}, \ie, assigning to each pixel a probability of being an object boundary \cite{Yang2016CEDN, DollarZ14}.

\begin{figure}[h!]
\centering
\includegraphics[width=.9\linewidth,page=1]{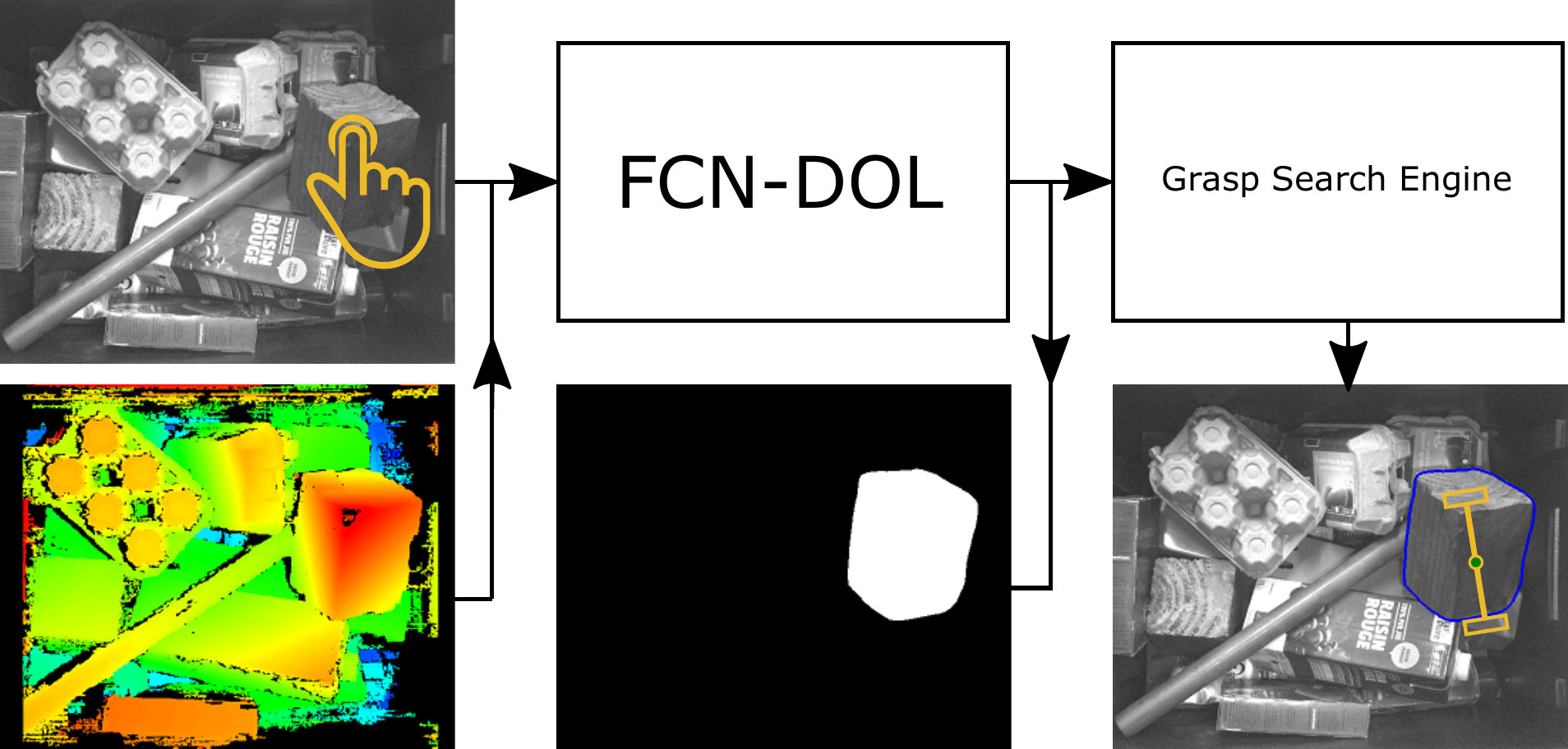}
\caption{A test application of our approach on a real robotic setup: (1) A user clicks on the object to extract. (2) The proposed model (FCN-DOL) delineates the selected object in a single forward pass. (3) An off-the-shelf grasp search engine can thus generate a relevant object-oriented robotic grasp.}
\label{fig:robotResults}
\end{figure}

\begin{figure}[h!]
\centering
\setlength{\mywidth}{.28\linewidth}
\subfloat[Image]{\includegraphics[width=\mywidth]{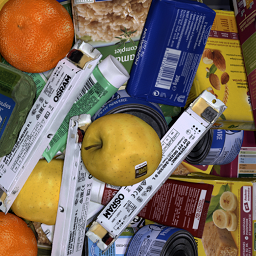}}\quad
\subfloat[Patch (box)]{\includegraphics[width=\mywidth]{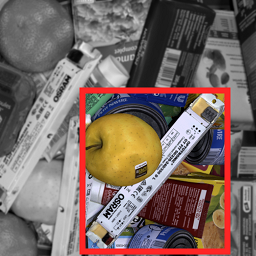}}\quad
\subfloat[Seed (point)]{\includegraphics[width=\mywidth]{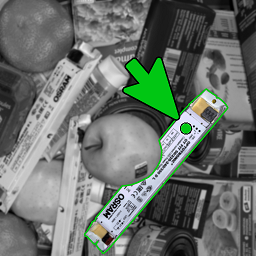}}
\caption{A use case showing that it is less ambiguous to consider the task ``Segment the object including this pixel'' (c) rather than ``Segment the object inside this box'' (b) as a patch may be equally shared by several objects.}
\label{fig:patchVsSeed}
\end{figure}

In this work, we aim at extracting one by one each instance of a dense pile of objects in random poses. As illustrated by Fig. \ref{fig:robotResults}, we consider the following scenario: (1) a depth-augmented image of the scene is captured; (2) a human operator clicks on the next object to extract; (3) the selected object is automatically delineated; (4) robotic grasps on the segmented object are automatically detected. This paper describes the third step, \ie, the segmentation of the selected object. As the ultimate goal is to successfully grasp each item of the scene, object instances must be correctly distinguished despite potential occlusion and entanglement, which is hardly achievable without an explicit notion of object. In the context of waste sorting, prior knowledge is not available: the target objects to be extracted, as well as their corresponding model, are unknown. Differently from industrial bin-picking of many instances of a known manufactured object, traditional pose estimation techniques \cite{ChoiTTLR12} cannot be applied. We thus adapt object segmentation methods used for everyday image analysis to the context of industrial robotic picking. Indeed, state-of-the-art techniques \cite{MaskRCNN, SharpMask} first perform a coarse localization by extracting a set of image patches assumed to contain one instance. While this approach is relevant for the typical foreground/background paradigm inherent to most indoor and outdoor scenes, it can be ambiguous when the scene is only composed of objects densely stacked on top of each other, as a patch may be equally shared by many objects (cf. Fig. \ref{fig:patchVsSeed}). The proposed approach alleviates this problem by considering a pixel included in the object, instead of a bounding box. Interestingly, one can notice in the present application context that inter-instance boundaries are complementary to instance segments, since the background is generally fully hidden and all objects are targets. This suggests besides that object contour detection may be used as a backbone for object segmentation. In the following, we refer to this property as \emph{edge-mask duality}.

Moreover, state-of-the-art learning-based edge detection and segmentation tasks require training datasets annotated on a per-pixel basis. Despite many efforts \cite{Firman_2016_CVPR_Workshops}, notably for real indoor scenes featuring household objects \cite{SilbermanECCV12, LaiBRF11}, and table-top sparse clutters \cite{RichtsfeldMPZV12}, none matches the camera point of view, neither the typical scene setup of encountered waste sorting applications. Although collecting adequate data is certainly not a problem given the recent spread of affordable RGB-D sensors, annotating reliably new images remains tedious, as current methods consist in time-consuming manual labeling. Real image datasets are therefore limited in terms of content, and costly to extend. This makes the availability of appropriate training data a factor hindering the widespread deployement of pixel-wise learning-based techniques in industrial environments.

In this paper, towards an interactive application for robotized waste sorting, we address the middle issue of learning object segmentation in depth maps, given one user click per object to segment. We aim at demonstrating that leveraging edge-mask duality with synthetic depth maps for end-to-end instance segmentation enables higher boundary precision than a patch-based approach, while ensuring a framework fully pluggable in an interactice interface. Starting from the observations that (1) realistic RGB-D images of synthetic scenes can be automatically generated by simulation, (2) learning from synthetic depth-augmented images has already shown promising results for semantic segmentation of indoor scenes \cite{Handa_2016_CVPR} and disparity estimation \cite{Mayer_2016_CVPR}, (3) fully convolutional networks trained end to end for object contour detection has proven very effective \cite{Yang2016CEDN}, our contribution is two-fold:

\begin{itemize}
\item \textbf{A synthetic expandable dataset for industrial robotic picking.} With custom scripts on top of Blender \cite{Blender}, we produce realistic rendered RGB-D images of object instances stacked in bulk, thus ensuring time-saving and error-free annotations used as pixel-wise ground truth during training.

\item \textbf{A scheme to train a fully convolutional edge detector for seed-based instance segmentation.} First, we show that object contour detectors can be trained on synthetic depth maps while performing almost equivalently on real data at inference time. Second, we introduce edge-mask duality with a novel dual-objective loss function to train a fully convolutional encoder-decoder in segmenting objects clicked interactively by a human operator. We show that the proposed approach, more suited for a practical human-robot interface, outperforms a state-of-the-art patch-based network when objects are piled up in bulk.
\end{itemize}

Our paper is organized as follows. After reviewing the state of the art in Section \ref{sec:state-of-the-art}, we describe the proposed training scheme in Section \ref{sec:duality} and our training data production pipeline in Section \ref{sec:synthesizing}. Then, Section \ref{sec:setup} details our experimental protocol. Results are discussed in Section \ref{sec:results}.

\section{Related Work}
\label{sec:state-of-the-art}

\subsection{Synthetic training data}

To our knowledge, learning from synthetic RGB-D images for pixel-wise object segmentation tasks has yet received little attention. In the context of generic object detection and recognition, a first kind of approach consists in generating synthetic images by projecting 3D models onto natural ``background'' images \cite{PengSAS15, RozantsevLF15}. However, these synthetic images contain by construction at most a few object instances that are implicitly assumed to be distinguishable from their environment, thus excluding many situations like the present case of dense stack of object instances. A second type of approach consists in simulating scenes from scratch. Mainly dedicated to semantic segmentation, synthetic datasets of annotated urban \cite{Ros_2016_CVPR} and indoor \cite{HandaPSC16} scenes have recently emerged. Following the creation of these datasets, it has been shown that depth-based semantic segmentation of indoor scenes with a convolutional neural network (CNN) can be boosted by using synthetic images as training data and ten times less real images only for fine-tuning \cite{Handa_2016_CVPR}. Also, in the footsteps of FlowNet \cite{DosovitskiyFIHH15} for optical flow estimation using synthetic images, training a CNN only on sufficiently realistic rendered image patches has again recently proved efficient for the purposes of disparity estimation on outdoor scenes with DispNet \cite{Mayer_2016_CVPR}. In continuity with the latter simulation-based approaches, we explore the context of robotic picking when the target objects are undefined. We show that real depth maps, whose costly pixel-wise ground-truth annotations are unsustainable in industrial environments, may be fully replaced by rendered depth maps when feeding structured random forests or fully convolutional networks for object contour detection during training.

\subsection{Instance segmentation}

Due to a lack of large real-world RGB-D datasets, instance segmentation algorithms have been mostly performed on RGB-only images of everyday scenes, augmented with crowd-sourced pixel-wise annotations \cite{LinMBHPRDZ14}. Concurrently with the emergence of end-to-end deep learning for instance segmentation, images were typically represented as a graph of superpixels, often built from boundary-related cues. To this end, popular edge detectors, using structured random forests to classify patches \cite{ZitnickD14} or lately end-to-end fully convolutional "encoder-decoder" networks \cite{Yang2016CEDN}, were employed to give edge-preserving starting cues. Unsupervised instance proposals generation then typically consisted in multi-scale combinatorial grouping \cite{Pont-TusetABMM15} based upon generic hand-crafted features, \eg, instance candidate's area or perimeter. Later, learning to propose instance candidates was notably introduced with conditional random fields (CRFs) \cite{KrahenbuhlK15}, that are jointly trained for binary graph cuts while maximizing global recall. Lately, Facebook AI Research released a network that predicts scored instance proposals from image patches \cite{PinheiroCD15}, trained end to end on the MS COCO dataset \cite{LinMBHPRDZ14}, and outperforming previous state-of-the-art approaches. A top-down refinement scheme was further proposed \cite{SharpMask} to refine the boundaries of the predicted masks, since without explicit notion of object contour during training, feed-forward deep architectures generally fail to encode accurate boundaries due to progressive down-sampling. Within the context of semantic instance segmentation, recurrent neural networks have been employed to sequentially segment each instance belonging to one inferred category \cite{RomeraParedes2016}, by heading fully convolutional networks for semantic segmentation with long short-term memory (LSTM) layers, so as to enable a recurrent spatial inhibition. In contrast to previous approaches, we leverage the duality between inter-instance boundaries and instance masks (described in Section \ref{sec:duality}) directly within a fast feed-forward architecture by introducing an explicit object contour learning during training for instance segmentation, so as to obtain masks with sharp and consistent boundaries in the case of objects piled up in bulk.

\section{Leveraging edge-mask duality}
\label{sec:duality}

\begin{figure}[b!]
\centering
\includegraphics[width=.9\linewidth]{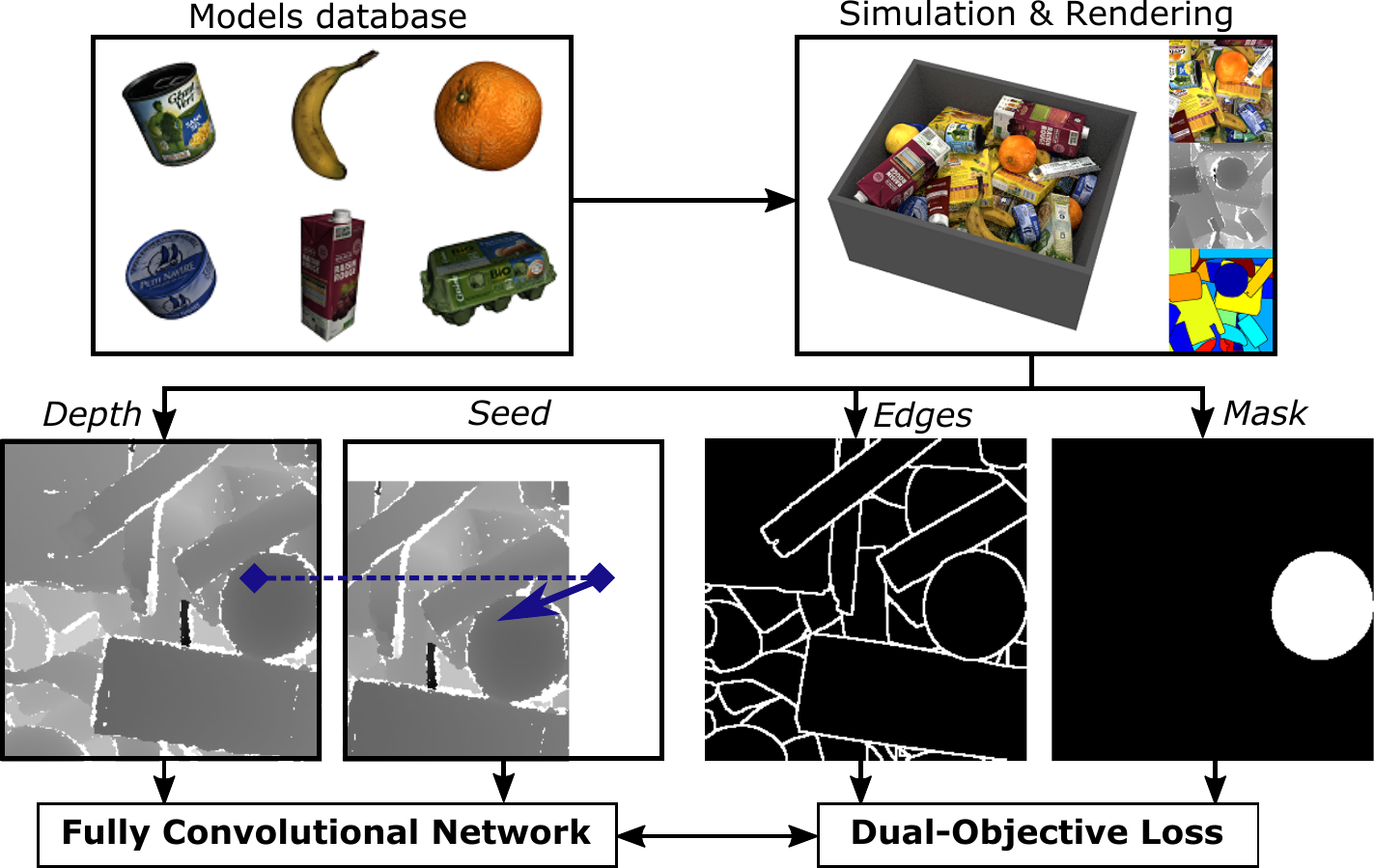}
\caption{Our training scheme to generate instance segments by introducing the notion of \emph{edge-mask duality} with a novel loss function. A fully convolutional encoder-decoder network is fed with one image and its translation centering the user click on one instance, namely seed, to predict the overall inter-instance boundary map and the instance mask including the input seed. The proposed dual-objective loss drives the network to explicitly learn segmentation from boundaries.}
\label{fig:trainingscheme}
\end{figure}

Motivated by the many recent works on feed-forward fully convolutional architectures that demonstrated strong learning capabilities and fast inference time to be part of a responsive interactive interface, we follow the idea that when the scene is full of target objects, typically in waste sorting applications, the inter-instance boundaries and the set of instance masks are dual representations, what we call \emph{edge-mask duality}. This interestingly suggests that feature encoding-decoding for object contour detection may be used as backbone for instance segmentation. One may indeed expect a better object pose invariance and a finer segmentation when learning local inter-instance boundaries instead of the spatial continuity inside the masks. We thus propose to repurpose a state-of-the-art encoder-decoder object contour detector to predict instance segments by appending during training to the input and the output respectively a seed pixel and the instance mask including the seed. State-of-the-art approaches \cite{MaskRCNN, SharpMask} link a patch, either randomly generated or predicted, to one instance, which requires size priors and may be ambiguous (cf. Fig. \ref{fig:patchVsSeed}). We use instead a translation centering a pixel location assumed to be included in one instance, namely seed, here provided by an external user, as illustrated by Fig. \ref{fig:trainingscheme}.

Formally, suppose we have a set of annotated depth maps. For one depth map, let $\mathcal{G}$ be the ground-truth instance label map and $\mathcal{E}$ the ground-truth inter-instance contour map, that is, for each pixel $\mathbf{p}$,
\begin{equation}
\mathcal{E}_\mathbf{p} = \left\{ 
\begin{aligned}
1&\quad  \text{if}\ \exists (\mathbf{q}, \mathbf{q'}) \in \mathcal{N}_\mathbf{p}^2 \text{ such that } \mathcal{G}_\mathbf{q}\neq \mathcal{G}_\mathbf{q'}\\
0 & \quad \text{otherwise}
\end{aligned} \right .
\end{equation}

\noindent where $\mathcal{N}_\mathbf{p}$ denotes the 8-connected neighborhood of pixel $\mathbf{p}$. In other words, $\mathcal{E}_\mathbf{p}$ is the expected probability that $\mathbf{p}$ is an instance boundary. Let $\{\mathcal{M}_k\}_{k}$ be the set of the corresponding ground-truth instance binary masks. One can simply define the instance mask $\mathcal{M}_\mathbf{s}$ at seed pixel $\mathbf{s}$ as a two-operation process on the contour map:
\begin{equation}\label{eq:duality}
\mathcal{M}_\mathbf{s} = \mathcal{C}(\bar{\mathcal{E}}, \mathbf{s})
\end{equation}

\noindent where $\mathcal{C}(\mathcal{X}, \mathbf{s})$ is the connected subset of $\mathcal{X}$ including $\mathbf{s}$, and $\bar{\mathcal{X}}$ the complementary of $\mathcal{X}$. $\mathcal{M}_\mathbf{s}$ thus defines the expected probability for each pixel $\mathbf{p}$ to belong to the instance including the seed $\mathbf{s}$. Reciprocally, the instance contour map can be trivially derived from the set of instance masks. 

Feed-forward networks require one-to-one or many-to-one relationships between the training inputs and expected outputs, however in the context of instance segmentation, the corresponding set of instances is of varying cardinality. Circumventing this issue thus requires each expected instance mask to be associated with one different input. State-of-the-art approaches \cite{SharpMask} implement one-to-one relationships by defining a set of patches, assuming that one patch contains one instance. In this work, we instead consider a set $\mathcal{S}$ of seed pixels and propose the following mapping:
\begin{equation}\label{eq:mapping}
\begin{aligned}
\quad \mathcal{S} & \to & \{\mathcal{M}_k\}_k\\
\mathbf{s} & \mapsto & \mathcal{M}_\mathbf{s}\hspace{.5cm}
\end{aligned}
\end{equation}

\noindent so that each seed pixel maps to the instance it belongs to. Exploiting the translational invariance of convolutional networks and Equation \ref{eq:mapping}, we can then define a pixel-wise training set to use encoder-decoder networks for instance segmentation, by mapping one depth map and one seed $\mathbf{s}$ to one contour map $\mathcal{E}$ and one instance mask $\mathcal{M}_\mathbf{s}$, as depicted by Figure \ref{fig:trainingscheme}. At training time, we select randomly a few seeds inside each ground-truth instance. At inference time, the user provides a seed by clicking on the instance.

To introduce the edge-mask duality in the learning, we train the encoder-decoder network end to end by minimizing the following sum of pixel-wise logistic loss functions
\begin{equation}
\mathcal{L}(\theta) = \frac{1}{N}\sum_{n=1}^{N} \left\{ \ell(\lambda_e, \mathcal{E}^{(n)}, \hat{\mathcal{E}}^{(n)}) + \sum_{\mathbf{s}\in \mathcal{S}} \ell(\lambda_m, \mathcal{M}_\mathbf{s}^{(n)}, \hat{\mathcal{M}}_\mathbf{s}^{(n)}) \right\}
\end{equation}

\noindent where $\theta$ denotes the network's parameters, $N$ the number of training samples, $\ell$ has the following form:

\begin{equation}
\begin{split}
\ell(\lambda, \mathcal{Y}, \hat{\mathcal{Y}}) = - \sum_\mathbf{p} \left\{ (1-\mathcal{Y}_\mathbf{p})\log(1 - \sigma(\hat{\mathcal{Y}}_\mathbf{p})) \right. \\
\left. +\ \ \ \lambda\ \ \mathcal{Y}_\mathbf{p}\log(\sigma(\hat{\mathcal{Y}}_\mathbf{p})) \right\}
\end{split}
\end{equation}

\noindent $\mathcal{Y}_\mathbf{p}\in\{0,1\}$ and $\hat{\mathcal{Y}}_\mathbf{p}\in\mathbb{R}$ are respectively the expected output and the encoder-decoder network's response at pixel $\mathbf{p}$, $\lambda$ is a trade-off parameter between "object contour" or "instance mask" pixels and inactive ones, and $\sigma$ is the sigmoid function. While the term $\ell(\lambda_e, \mathcal{E}, \hat{\mathcal{E}})$ focuses the network's attention on boundaries, the term $\ell(\lambda_m, \mathcal{M}_\mathbf{s}, \hat{\mathcal{M}}_\mathbf{s})$ enables to learn per-seed connectivity, so that encoding-decoding for predicting instance contours on the whole image and instance masks is jointly learned, as suggested by Equation \ref{eq:duality}. In practice, we set $\lambda_e=10$ and $\lambda_m=1$. By definition, $\mathcal{L}$ is fully differentiable as a linear combination of logistic losses. With the proposed loss function, instance segmentation is performed in one forward pass while learning explicitly driven by inter-instance boundaries.

\section{Synthesizing training images}
\label{sec:synthesizing}

To train the network, we generate a synthetic dataset using Blender \cite{Blender} with custom code to simulate scenes of objects in bulk and render the corresponding RGB-D top views.

\begin{figure}[h]
\centering
\setlength{\mywidth}{.17\linewidth}

\subfloat{\includegraphics[width=\mywidth]{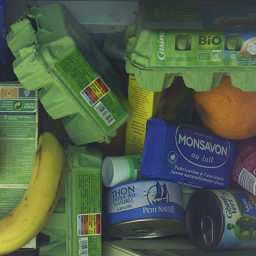}\quad
\includegraphics[width=\mywidth]{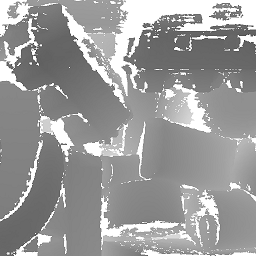}\quad
\includegraphics[width=\mywidth]{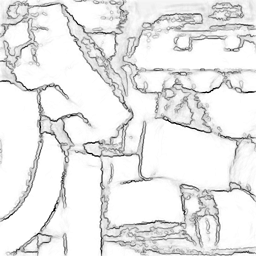}\quad
\includegraphics[width=\mywidth]{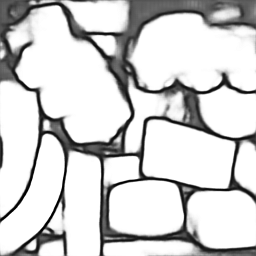}\quad
\includegraphics[width=\mywidth]{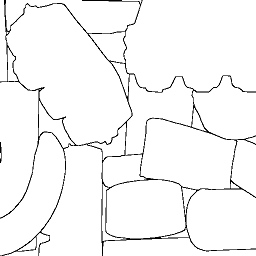}}

\subfloat{\includegraphics[width=\mywidth]{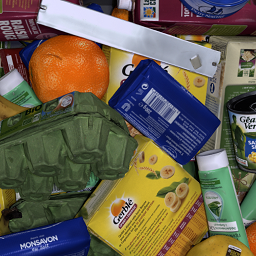}\quad
\includegraphics[width=\mywidth]{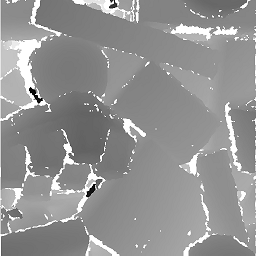}\quad
\includegraphics[width=\mywidth]{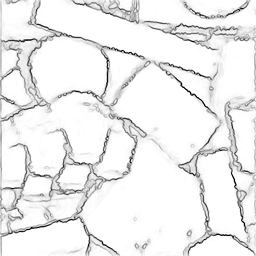}\quad
\includegraphics[width=\mywidth]{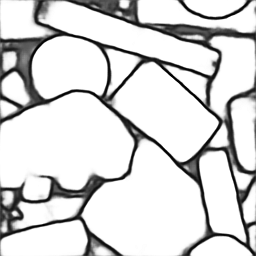}\quad
\includegraphics[width=\mywidth]{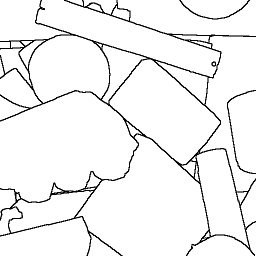}}

\caption{Detected object contours on real (first row) and synthetic (second row) test images, both using the same model trained only on our synthetic depth maps. From left to right: RGB, depth, contours using a structured random forest (RF) \cite{DollarZ14}, contours using a fully convolutional network (FCN) \cite{Yang2016CEDN}, ground truth.}
\label{fig:edgedetection}
\end{figure}

\paragraph{Simulation}

We model a static bin, an RGB-D sensor and some objects successively dropped above the bin in random pose using Blender's physics engine. We add some intraclass geometric variability if needed, by applying isotropic and anisotropic random scaling. In both our real robotic setup and Blender, depth is recovered using an active binocular stereoscopic system composed of two RGB cameras and a pseudo-random pattern projector to add artificial texture to the scene. Calibration, rectification and matching are performed using standard off-the-shelf algorithms.

\paragraph{Rendering}

Once all instances have been dropped, we render each camera view using Cycles render engine and process the pair of rendered images in the exact same way as a pair of real images. The depth sensor's noise model is thus implicitly embedded during stereo matching, which gives realistic depth maps without additional post-processing, contrary to \cite{Handa_2016_CVPR}. Figure \ref{fig:edgedetection} shows a qualitative comparison between real and synthetic images from our dataset.

\section{Experimental setup}
\label{sec:setup}

We conduct two sets of experiments: (1) we train two state-of-the-art object contour detectors -- a structured random forest (RF) \cite{DollarZ14} and a fully convolutional network (FCN) \cite{Yang2016CEDN} -- on synthetic depth maps to achieve equivalent performance on real data at inference time; (2) we show that learning instance segmentation by training a fully convolutional encoder-decoder network with our novel dual-objective loss function (FCN-DOL) outperforms a patch-based network (DeepMask) \cite{SharpMask} in terms of boundary precision, while requiring only seed pixels instead of patches. In all our experiments, we focus on synthetic depth as rendered RGB is less realistic and depth conveys more generalizable information.

\paragraph{Data preparation}

To evaluate our synthetic training, we capture 25 RGB-D top views of real-world scenes featuring 15 household objects. Four expert human annotators manually label each image with the ground-truth contours. Using an off-the-shelf 3D scanner, we scan each of the 15 objects, and generate 1,000 synthetic images of multi-object scenes, \ie, scenes with many instances of many objects, and 1,500 mono-object scenes (100 per object), \ie, scenes with many instances of one object. The multi-object scenes are randomly divided into 3 subsets with the following distribution: 400 for training, 200 for validation, and 400 for test. All mono-object scenes are used only for test. In addition, 5 synthetic and 5 real images are manually labeled separately, each one by 3 different annotators, to evaluate the quality of manual annotation and the impact of ground truth on contour detection performances. We limit the manual annotation to a few images due to the burden of the task, that required 68 hours in total.

\paragraph{Performance metrics}

For contour detection, we use standard metrics from \cite{Arbelaez2011}: the best F-score, and the corresponding recall and precision, with a fixed threshold on dataset scale. We add the \emph{synthetic gap}, \ie the absolute difference between scores on real (R) and synthetic (S) test images. For instance segmentation, we compute two quantities: the average intersection over union ratio (IoU) over the best ground-truth-matching proposal set; and the boundary precision on instance scale using the same metric as for contour detection. Given one image, a proposal best matches a ground-truth instance if it has the highest IoU among all proposals for one of the ground-truth instances, and if the corresponding IoU is higher than 50\%.

\paragraph{Settings for evaluating the use of synthetic images}

We train two object contour detectors, state-of-the-art in their kind: a structured random forest (RF) from \cite{DollarZ14} of 4 trees and a fully convolutional network (FCN) from \cite{Yang2016CEDN} with a VGG-16-based encoder-decoder architecture -- on synthetic depth maps to perform equivalently on real data at inference time. For RF, the trees are trained on $10^6$ patches extracted from 100 synthetic multi-object images ($10^4$ patches -- as many positive as negative ones -- per training image). Multi-scale option and non-maximal suppression are not enabled at test time. Other parameters are set to their default value. For FCN, we perform 180 epochs on 400 synthetic images, using Caffe \cite{jia2014caffe}. Depth is introduced as a new input layer replacing the BGR inputs, therefore both encoder and decoder's weights are learned. Weights are initialized like in \cite{Yang2016CEDN}, except that two channels of the input layer are removed. We set a constant learning rate of $10^{-4}$ and a weight decay of $10^{-4}$. For both RF and FCN, tests are performed on the 25 manually annotated real depth maps (R-tests), and on 25 synthetic depth maps (S-tests) of multi-object scenes. Human annotators are evaluated on 5 real and 5 synthetic images.

\paragraph{Settings for evaluating the use of edge-mask duality}

We train two feed-forward deep architectures only on synthetic depth maps of multi-object scenes: (i) the state-of-the-art patch-based network referred to as DeepMask \cite{SharpMask}, built on a 50-layer residual encoder \cite{HeZRS16} and taking patches as input; (ii) the previous encoder-decoder network referred to as FCN \cite{Yang2016CEDN} but trained using our dual-objective loss function (FCN-DOL). For these trainings, we removed all partially occluded instances from ground truth, so that each model is trained only on instances whose mask is connected and of reasonably large area, \ie, graspable instances. DeepMask is initialized with the weights pretrained on MS COCO, and fine-tuned by performing 100 epochs on our 400 synthetic training images, with a constant learning rate of $10^{-4}$ and data augmentation (shift and scale jittering). FCN-DOL is initialized with our FCN model for contour detection. We add one input layer, corresponding to the translated image that centers the seed pixel, and one output layer, corresponding the object mask including the seed. We then perform 120 epochs driven by our dual-objective loss function, with a constant learning rate of $10^{-4}$, using 2 random pixel seeds per ground-truth instance and data augmentation (4 rotations and horizontal mirror). At test time, we define a constant regular grid of seeds over the whole image. For each image, we obtain a set of object mask proposals by forwarding each seed of the grid, and binarizing the continuous output mask with a threshold of 0.8. For both DeepMask and FCN-DOL, tests are performed on 400 synthetic depth maps of multi-object scenes and 1,500 synthetic images of mono-object scenes.

\section{Results}
\label{sec:results}

\paragraph{Synthetic over real training images}

Using simulation enables significant time-savings: annotating a real image takes 40 min in average while generating a synthetic view about 5 min (with a quad-core 3.5GHz Intel Xeon E5 for physics simulation, and a Nvidia Quadro M4000 for GPU rendering), thus 8 times faster. Interestingly, both edge detectors trained on synthetic depth maps still achieve good results in practice when tested on our real data, producing contour maps visually similar to those obtained with a synthetic input, as illustrated in Fig. \ref{fig:edgedetection}. However, Tab. \ref{tab:edgedetection} shows synthetic gaps of 16 and 18 points in F-score inducing that our synthetic data is not realistic enough, here in terms of depth noise modeling. Nevertheless, this gap can be partially relativised, considering the non-zero gap (12 points) achieved by humans. S-tests are indeed conducted against perfect ground truth while R-tests against error-prone annotations, as humans miss hardly detectable edges in noisy areas. Bridging the gap with more realism should be part of a future work to limit any overtraining in the synthetic domain.

\begin{table}[ht]
\centering
\begin{tabu}{l|[1.5pt]c|[1.5pt]c|[1.5pt]c|c|c|[1.5pt]c|c|c|[1.5pt]c|c|c}
\tabucline[1.5pt]{4-}
\multicolumn{3}{c}{}
&\multicolumn{9}{c}{Test}\\
\tabucline[1.5pt]{4-}
\multicolumn{3}{c}{}
&\multicolumn{3}{c|[1.5pt]}{Recall}
&\multicolumn{3}{c|[1.5pt]}{Precision}
&\multicolumn{3}{c}{F-score}\\
\tabucline[1.5pt]{2-}
\multicolumn{1}{c}{}
&Modality
&Train
&25R&25S&\textbf{Gap}
&25R&25S&\textbf{Gap}
&25R&25S&\textbf{Gap}\\
\tabucline[1.5pt]{-}
RF \cite{DollarZ14}
&Depth
&100S
&.54&.70&\textbf{.16}
&.53&.69&\textbf{.16}
&.53&.69&\textbf{.16}\\
\hline
FCN \cite{Yang2016CEDN}
&Depth
&400S
&.43&.60&\textbf{.17}
&.78&.93&\textbf{.15}
&.55&.73&\textbf{.18}\\
\hline
Humans
&RGB
&-
&.48&.65&\textbf{.17}
&.98&.92&\textbf{.06}
&.64&.76&\textbf{.12}\\
\tabucline[1.5pt]{-}
\end{tabu}
\caption{Comparative results for instance contour detection on real (R) and synthetic (S) images, using a random forest (RF) \cite{DollarZ14} and a fully convolutional network (FCN) \cite{Yang2016CEDN} both trained only on our synthetic depth maps. The synthetic gap (Gap) is the absolute difference between R and S-scores. Humans are given RGB since they are unable to distinguish instances only from depth.}
\label{tab:edgedetection}
\end{table}

\begin{table}[h!]
\centering
\begin{tabu}{l|[1.5pt]c|[1.5pt]c|c|[1.5pt]c|c}
\tabucline[1.5pt]{3-}
\multicolumn{2}{c}{}&\multicolumn{2}{c|[1.5pt]}{400Mu}&\multicolumn{2}{c}{1500Mo}\\
\tabucline[1.5pt]{2-}
\multicolumn{1}{c}{}&\multirow{2}*{Train}&Average&Boundary&Average&Boundary\\
\multicolumn{1}{c}{}&&Best IoU&Precision&Best IoU&Precision\\
\tabucline[1.5pt]{-}
DeepMask \cite{SharpMask}&400Mu&.83&.51&.88&.33\\
\tabucline[1.5pt]{-}
FCN-DOL (Ours)&400Mu&\textbf{.89}&\textbf{.90}&\textbf{.88}&\textbf{.90}\\
\tabucline[1.5pt]{-}
\end{tabu}
\caption{Intersection over Union (IoU) and boundary precision of the best matching instances on synthetic multi-object (Mu) and mono-object (Mo) scenes.}
\label{tab:instancedetection}
\end{table}

\begin{figure}[h!]
\centering
\setlength{\mywidth}{.23\linewidth}

\subfloat{\includegraphics[width=\mywidth]{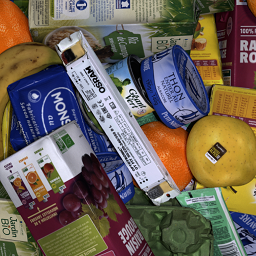}}\qquad
\subfloat{\includegraphics[width=\mywidth]{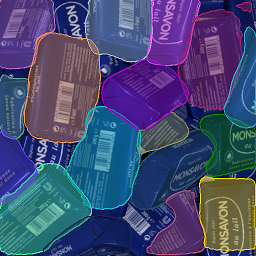}}\hspace{0.05cm}
\subfloat{\includegraphics[width=\mywidth]{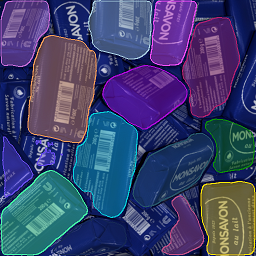}}\hspace{0.05cm}
\subfloat{\includegraphics[width=\mywidth]{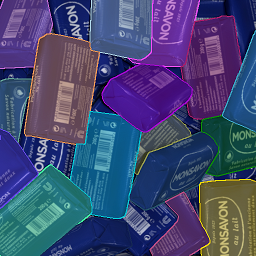}}

\subfloat{\includegraphics[width=\mywidth]{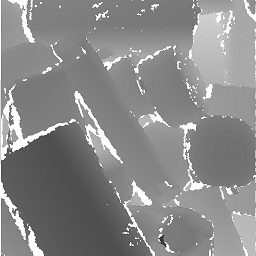}}\qquad
\subfloat{\includegraphics[width=\mywidth]{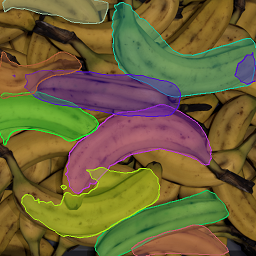}}\hspace{0.05cm}
\subfloat{\includegraphics[width=\mywidth]{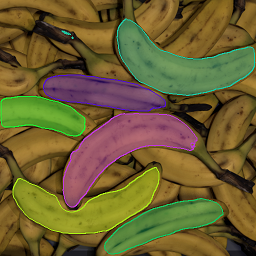}}\hspace{0.05cm}
\subfloat{\includegraphics[width=\mywidth]{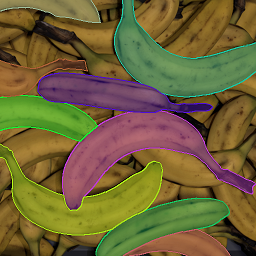}}

\addtocounter{subfigure}{-8}

\subfloat[Training sample]{\includegraphics[width=\mywidth]{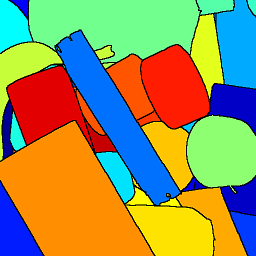}}\qquad
\subfloat[DeepMask \cite{SharpMask}]{\includegraphics[width=\mywidth]{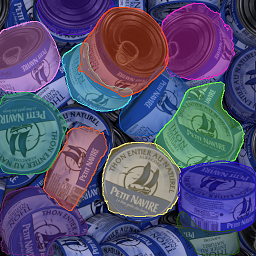}}\hspace{0.05cm}
\subfloat[\fontsize{8pt}{9pt}\selectfont FCN-DOL (Ours)]{\includegraphics[width=\mywidth]{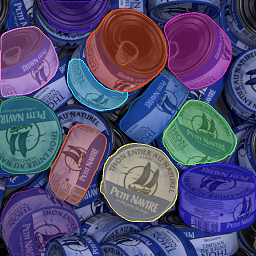}}\hspace{0.05cm}
\subfloat[Ground truth]{\includegraphics[width=\mywidth]{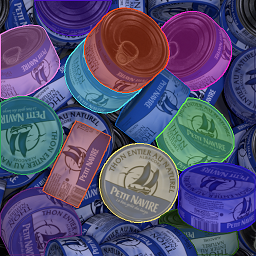}}

\caption{Best matching instance proposals on synthetic test depth maps. With one depth-only training on multi-object scenes (a), our edge-driven model (c) gives more edge-preserving instances and better generalizes to unknown arrangements, in contrast to patch-based networks (b).}
\label{fig:results}
\end{figure}

\paragraph{Seed-based edge-driven over patch-based instance segments}

As shown by Fig. \ref{fig:results}, our dual-objective loss for instance segmentation by learning explicitly inter-object boundaries leads to sharper mask boundaries than using a state-of-the-art patch-based network, when the scene is a pile of bulk objects as it is the case in a waste sorting application. These results are corroborated by Tab. \ref{tab:instancedetection}, with a gain of 39 points in boundary precision. In addition, our model performs equally on mono-object setups though it was trained only on multi-object scenes, whereas the patch-based network's performance drops by 18 points. This suggests that introducing the edge-mask duality at training time enables a better generalization on new arrangements. In a real interactive interface, the human-guided network helps to relevantly reduce the grasp search space as illustrated by Fig. \ref{fig:robotResults}. Although the proposed model implies one forward pass per seed, this is not problematic in our application context as we trust the user to perform a relevant click. Our experimental dataset should nevertheless be extended to a much larger database of 3D models such as ShapeNet \cite{ChangFGHHLSSSSX15} to investigate the generalization on new objects, as our current images lack variability in terms of object category.

\paragraph{Limitations of the proposed model and future work}

\begin{figure}[h!]
\centering
\setlength{\mywidth}{.15\linewidth}

\subfloat{\includegraphics[width=\mywidth]{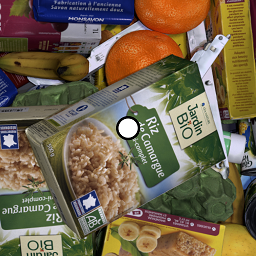}}\hspace{0.05cm}
\subfloat{\includegraphics[width=\mywidth]{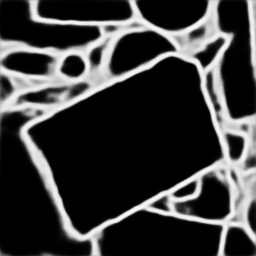}}\hspace{0.05cm}
\subfloat{\includegraphics[width=\mywidth]{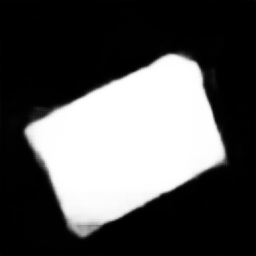}}\hspace{0.05cm}
\subfloat{\includegraphics[width=\mywidth]{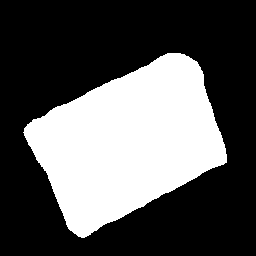}}\hspace{0.05cm}
\subfloat{\includegraphics[width=\mywidth]{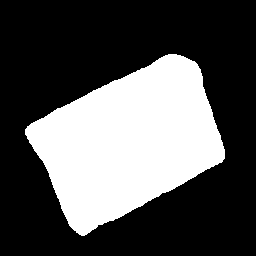}}\hspace{0.05cm}
\subfloat{\includegraphics[width=\mywidth]{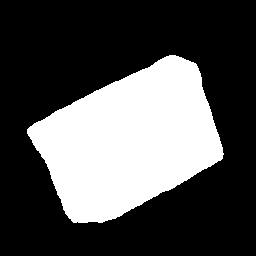}}

\subfloat{\includegraphics[width=\mywidth]{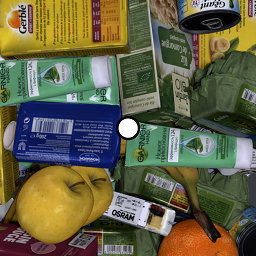}}\hspace{0.05cm}
\subfloat{\includegraphics[width=\mywidth]{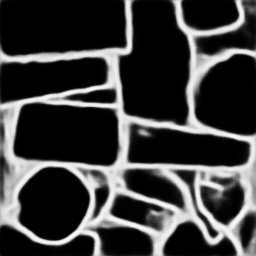}}\hspace{0.05cm}
\subfloat{\includegraphics[width=\mywidth]{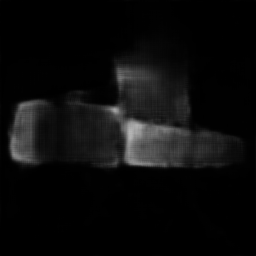}}\hspace{0.05cm}
\subfloat{\includegraphics[width=\mywidth]{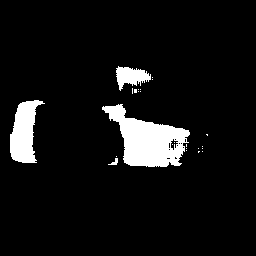}}\hspace{0.05cm}
\subfloat{\includegraphics[width=\mywidth]{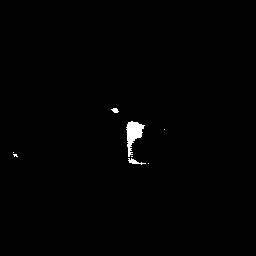}}\hspace{0.05cm}
\subfloat{\includegraphics[width=\mywidth]{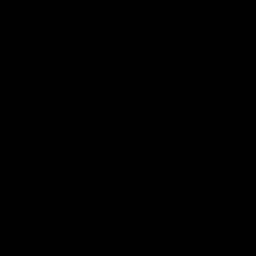}}

\subfloat{\includegraphics[width=\mywidth]{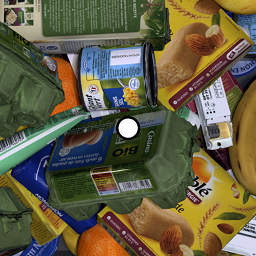}}\hspace{0.05cm}
\subfloat{\includegraphics[width=\mywidth]{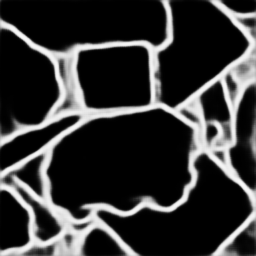}}\hspace{0.05cm}
\subfloat{\includegraphics[width=\mywidth]{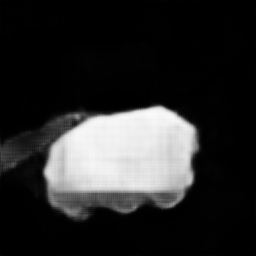}}\hspace{0.05cm}
\subfloat{\includegraphics[width=\mywidth]{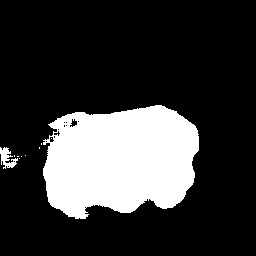}}\hspace{0.05cm}
\subfloat{\includegraphics[width=\mywidth]{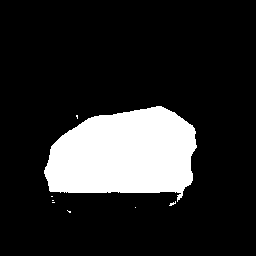}}\hspace{0.05cm}
\subfloat{\includegraphics[width=\mywidth]{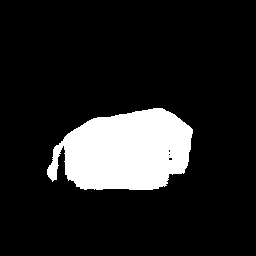}}

\caption{Examples of clicks (the white disks) using the proposed model. From left to right: RGB, predicted edges, predicted continuous mask, binarized masks with respectively a threshold of 0.2, 0.5, 0.8. From top to bottom: a click near the instance's centroid; a click on a boundary; a click near a boundary.}
\label{fig:threshold}
\end{figure}

The proposed model relies on a state-of-the-art encoder-decoder structure with 130M parameters against less than 60M parameters for DeepMask. This, and the lack of residual blocks, hinder the training, which explains why we performed a two-phase training with three times more fine-tuning epochs than DeepMask in total to reach a stable accuracy. Building a new encoder-decoder architecture with residual blocks and less filters for the layer connecting the encoder and the decoder -- currently 79\% of the parameters to train -- could alleviate this issue. Also, thanks to the translation invariance of fully convolutional networks, any pixel belonging to the object may be selected, and in practice, we can expect a reasonable seed, as a human user instinctively clicks near the object's centroid. As depicted in Fig. \ref{fig:threshold}, a click on a boundary produces a low-contrasted output as the selected seed doesn't belong to any instance. A click near a boundary returns a continuous segmentation map more sensitive to the binarization threshold. The robustness on this point could be increased with a smarter seed selection at training time.

\section{Conclusion}

We presented a simulation-based framework towards an interactive application for robotized waste sorting. Realistic synthetic RGB-D images are generated to train an encoder-decoder network for edge-driven instance segmentation in depth maps. Assuming edge-mask duality in the images, we introduce a novel dual-objective loss function to explicitly focus the learning on inter-instance boundaries. Given seeds provided interactively by a human operator, the network thus produces instance masks with sharper boundaries than a state-of-the-art patch-based approach, in a single forward pass, and better generalizes to unknown object arrangements. Its deep fully convolutional architecture enables strong learning capabilities into a responsive human-robot interface to speed up the overall process of selecting, extracting and sorting objects.

\bigskip

\textbf{Note:} This is a pre-print of an article published in Human Friendly Robotics, Springer Proceedings in Advanced Robotics, vol 7. The final authenticated version is available online at: \url{https://doi.org/10.1007/978-3-319-89327-3_16}


\bibliographystyle{abbrv}
\bibliography{mybibliography}   

%
%
%

\end{document}